\pdfoutput=1

\documentclass[letterpaper, 10 pt, journal, twoside]{ieeetran}
\usepackage{amsmath,amsfonts}
\usepackage{algorithm}
\usepackage{algpseudocode}

\usepackage{array}
\usepackage[caption=false,font=normalsize,labelfont=sf,textfont=sf]{subfig}
\usepackage{textcomp}
\usepackage{url}
\usepackage{verbatim}
\usepackage{graphicx}
\usepackage{cite}
\usepackage{xcolor,soul}
\usepackage{graphicx}
\usepackage{amsmath}
\usepackage{cite}
\usepackage{url}

\usepackage{times}
\usepackage{fancyhdr,graphicx,amsmath,amssymb}

\usepackage{algorithm}
\usepackage{algpseudocode}
\usepackage{soul}

\usepackage[font=footnotesize,belowskip=-0.5pt,aboveskip=0pt]{caption}
\usepackage{bigstrut}
\usepackage[utf8]{inputenc}
\usepackage{array}
\usepackage{multirow}
\usepackage{tabularx}
\usepackage{bigstrut}
\usepackage{makecell}
\usepackage{tabu}
\usepackage{bm}
\usepackage{caption}

\usepackage{amsmath}          
\usepackage{graphics}
\usepackage{graphicx}
\usepackage{float}
\usepackage{caption}
\usepackage{subcaption}
\usepackage{dblfloatfix}
\usepackage{calc}
\usepackage{url}
\usepackage{textcomp}
\usepackage{balance}
\usepackage{siunitx}
\usepackage{mdwlist}
\usepackage{comment}
\usepackage{enumerate}
\usepackage{lipsum}

\usepackage{gensymb}

\usepackage{array}
\usepackage{multirow}
\usepackage{tabularx}
\usepackage{booktabs}

\usepackage{bigstrut}
\usepackage{fixltx2e}

\usepackage{color}
\usepackage{hhline}
\usepackage[dvipsnames]{xcolor}
\usepackage{float}
\usepackage{caption}
\usepackage{balance}
\usepackage[colorlinks=true, linkcolor=blue, citecolor=blue, urlcolor=blue]{hyperref}
\algnewcommand{\LeftComment}[1]{\Statex\(\triangleright\)~\textcolor{blue}{#1}}

\begin{document}

\title{A Hyperspectral Imaging Guided Robotic Grasping System}
\author{Zheng Sun$^{1}$, Zhipeng Dong$^{1}$, Shixiong Wang$^{1}$, Zhongyi Chu$^{2}$,~\IEEEmembership{Member,~IEEE,} \\and~Fei~Chen $^{\dag 1}$,~\IEEEmembership{Senior~Member,~IEEE}
\thanks{Manuscript received: January, 23, 2025; Revised March, 24, 2025; Accepted May, 26, 2025.}
\thanks{This paper was recommended for publication by Editor Júlia Borràs Sol upon evaluation of the Associate Editor and Reviewers' comments. This study was supported in part by the InnoHK initiative of the Innovation and Technology Commission of the Hong Kong Special Administrative Region Government via the Hong Kong Centre for Logistics Robotics.}
\thanks{
$^{1}$ Zheng Sun, Zhipeng Dong, Shixiong Wang, and Fei Chen are with the Department of Mechanical and Automation Engineering, T-Stone Robotics Institute, The Chinese University of Hong Kong, Hong Kong SAR {\tt \footnotesize zhengsun@link.cuhk.edu.hk, zhipengdong@cuhk.edu.hk,
sxwang@hklcr.hk, 
f.chen@ieee.org}.
}
\thanks{$^{2}$ Zhongyi Chu is with the School of Instrumentation and Optoelectronic Engineering, Beihang University, Beijing 100191, China  {\tt \footnotesize chuzy@buaa.edu}.}
\thanks{
$^{\dag}$ Corresponding authors
}
\thanks{Digital Object Identifier (DOI): see top of this page.}
}

\markboth{IEEE Robotics and Automation Letters. Preprint Version. May, 2025}
{SUN \MakeLowercase{\textit{et al.}}: A Hyperspectral Imaging Guided Robotic Grasping System} 

\maketitle
\begin{abstract}
Hyperspectral imaging is an advanced technique for precisely identifying and analyzing materials or objects. However, its integration with robotic grasping systems has so far been explored due to the deployment complexities and prohibitive costs. Within this paper, we introduce a novel hyperspectral imaging-guided robotic grasping system. The system consists of PRISM (Polyhedral Reflective Imaging Scanning Mechanism) and the SpectralGrasp framework. PRISM is designed to enable high-precision, distortion-free hyperspectral imaging while simplifying system integration and costs. SpectralGrasp generates robotic grasping strategies by effectively leveraging both the spatial and spectral information from hyperspectral images. The proposed system demonstrates substantial improvements in both textile recognition compared to human performance and sorting success rate compared to RGB-based methods. Additionally, a series of comparative experiments further validates the effectiveness of our system. The study highlights the potential benefits of integrating hyperspectral imaging with robotic grasping systems, showcasing enhanced recognition and grasping capabilities in complex and dynamic environments. The project is available at: \href{https://zainzh.github.io/PRISM}{https://zainzh.github.io/PRISM}.

\end{abstract}


\section{Introduction}
\label{section:: introduction}
\IEEEPARstart{H}{yperspectral} imaging offers unparalleled capabilities for precise material identification and analysis by capturing rich spectral information across dozens to hundreds of contiguous bands \cite{lodhi2019hyperspectral}. This technology has been widely applied in diverse fields, including remote sensing \cite{bioucas2013hyperspectral}, biochemical analysis \cite{silva2018hyperspectral}, and industrial sorting \cite{tatzer2005industrial}. Recognizing its potential, researchers have recently begun exploring its integration into robotics to enhance material recognition and environmental perception \cite{erickson2020multimodal}. Such advancements are critical for enabling robotic systems to perform complex grasping and decision-making manipulation tasks.

Despite the potential advantages, the adoption of hyperspectral imaging in robotics remains limited. First, existing linescan hyperspectral systems typically necessitate auxiliary mechanisms, such as conveyor belts, rendering them more suitable for industrial-scale applications than for close-range robotic manipulation tasks \cite{elmasry2010principles,zheng2018discrimination}. Meanwhile, commercially available snapshot hyperspectral cameras are often twice as expensive as linescan cameras and offer fewer spectral bands, thus constraining their practical deployment in robotic systems \cite{thomas2025trends}. Furthermore, the inherent high-dimensionality of hyperspectral data introduces significant computational and storage challenges, hindering real-time application in dynamic environments \cite{makarenko2022real}. Lastly, while conventional hyperspectral applications emphasize spectral analysis, the equally valuable spatial information is frequently overlooked. Effective exploitation of both spectral and spatial information could significantly enhance robotic perception and manipulation performance.

\begin{figure}[tbp!]
\centering
	\includegraphics[width=0.92\linewidth]{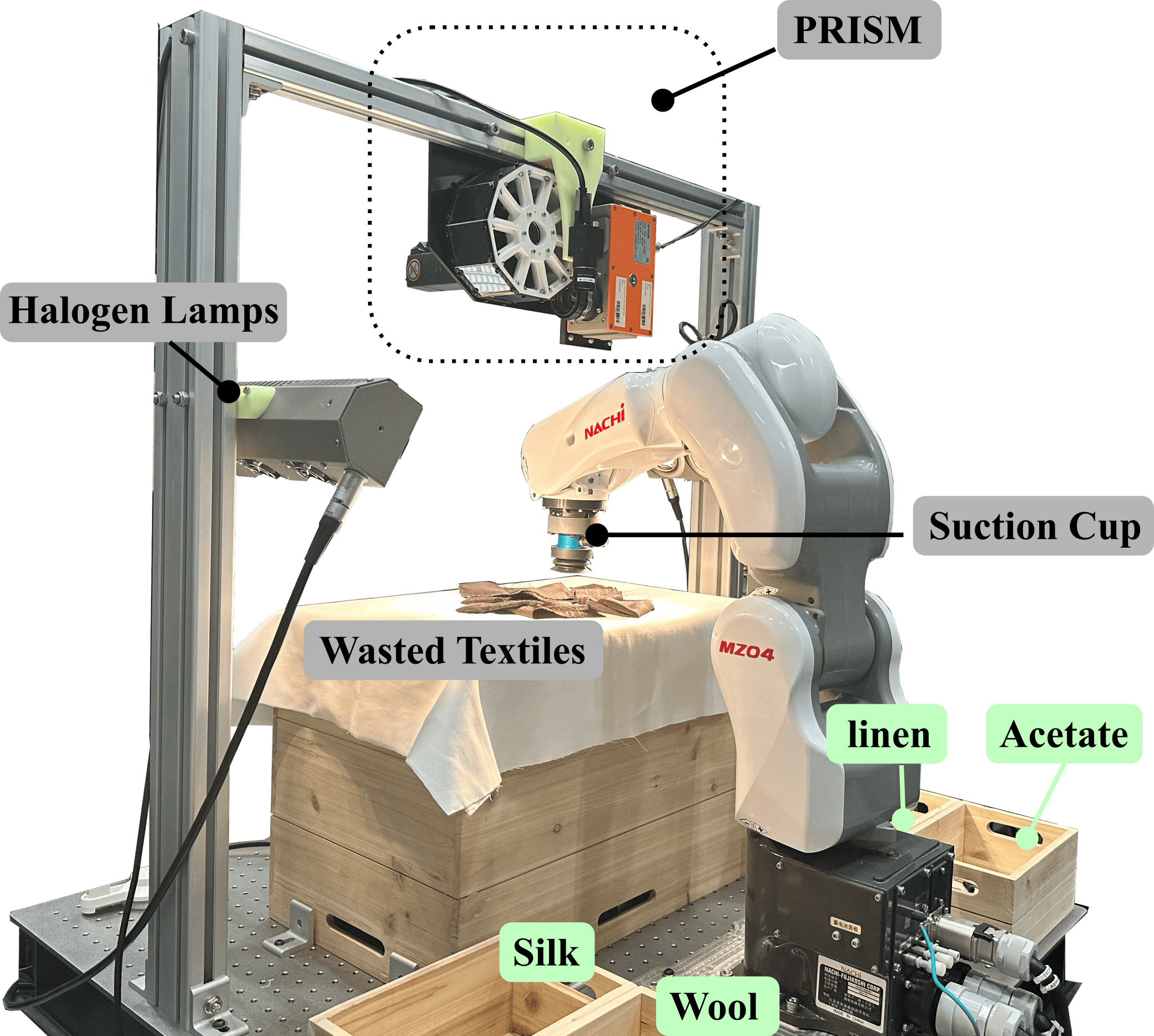}
        \vspace{5pt}
	\caption{Tabletop setup for the hyperspectral textile robotic grasping system. The setup includes a PRISM for hyperspectral image acquisition, halogen lamps, a Bernoulli's Principle suction cup gripper attached to a robotic arm, and a set of wasted textiles (linen, acetate, silk, wool) for sorting.}
    \label{fig:: first figure}
\end{figure}

To address these challenges, this paper introduces a hyperspectral imaging-guided robotic grasping system that effectively integrates hyperspectral perception with robotic grasping tasks. The proposed system consists of two core components: PRISM, a compact, cost-effective hyperspectral imaging device providing high precision and distortion-free hyperspectral imaging, which was first introduced in \cite{sun2024prism}; and SpectralGrasp, a novel framework that exploits hyperspectral imagery to generate precise and effective grasping strategies. To the best of our knowledge, few studies have focused on hyperspectral imaging-guided robotic grasping. Our system effectively integrates hyperspectral data into close-range robotic grasping and manipulation tasks, as illustrated in Fig. \ref{fig:: first figure}. The effectiveness of the proposed system is validated through textile sorting experiments, demonstrating superior recognition accuracy and sorting performance compared to human operators and conventional RGB-based methods. This study highlights the potential of hyperspectral imaging for enhancing robotic perception capabilities. This work makes the following contributions:
\begin{itemize}
    \item [1)] The design and implementation of PRISM, a hyperspectral imaging device achieving high-precision and distortion-free hyperspectral imaging with simplified integration into robotic systems.

    \item [2)] The development of SpectralGrasp, a framework integrating spectral and spatial information from hyperspectral images to generate effective robotic grasping strategies.

    \item [3)] Comprehensive validation through comparative experimental studies demonstrating improved recognition accuracy and sorting success rates relative to human performance and traditional RGB-based approaches.
\end{itemize}

\begin{figure*}[t]
	\includegraphics[width=1\linewidth]{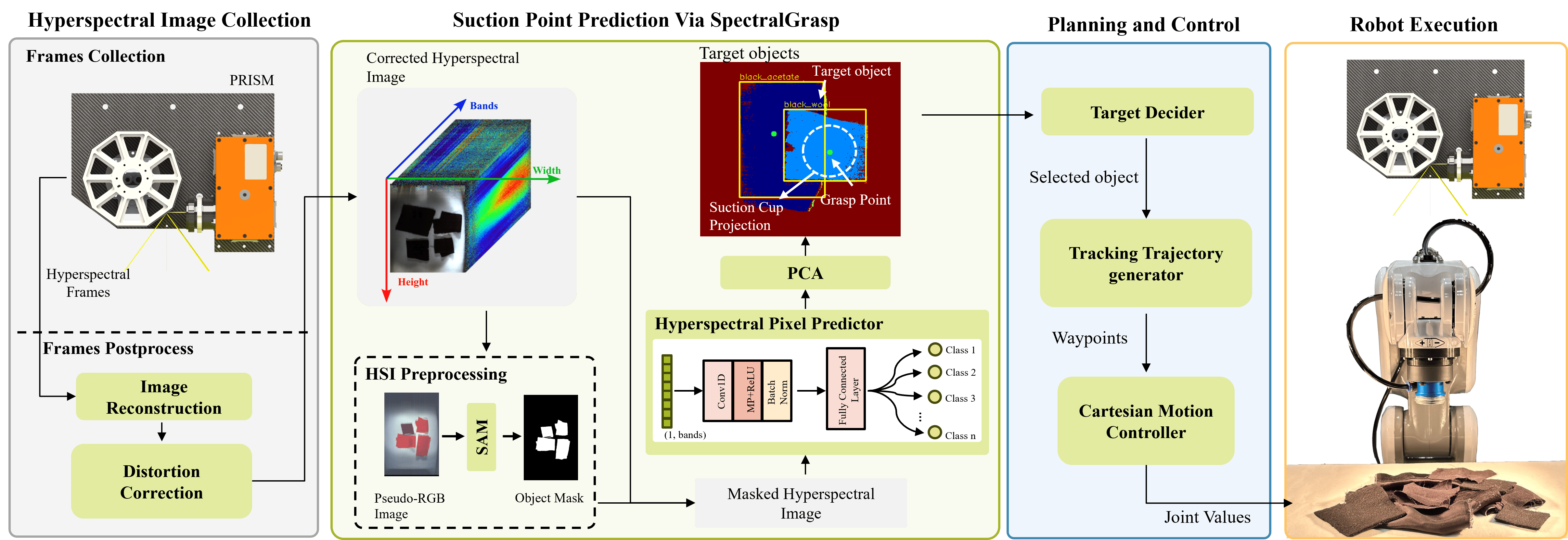}
        \vspace{1pt}
	\caption{System Overview: SpectralGrasp leverages both the spectral and spatial information from hyperspectral images captured by PRISM to generate suction points, enabling robots to complete textile sorting tasks. The process begins with PRISM capturing hyperspectral image frames, which are processed through image reconstruction and distortion correction to produce a corrected hyperspectral image. This corrected image with its mask serves as the input for SpectralGrasp. A pixel-level hyperspectral classifier is applied to generate pixel-level classification results. These results are then aggregated into object-level recognition outcomes using a Principal Component Analysis (PCA) algorithm. The system subsequently identifies the geometric centroids of target objects and generates a series of suction points for each target. A Target Decider module selects the target object, based on the task requirements, to generate the robotic execution trajectory. The trajectory is then sent to the Cartesian Motion Controller for precise sorting and execution.}
    \label{fig: System Overview}
\end{figure*}

\section{Related Works}

\subsection{Robotics Sensing Technologies}
Modern robotic manipulation heavily relies on perception, particularly in robotic grasp tasks. Early research primarily relied on RGB vision systems to capture an object's external shape and basic visual features such as color and  texture\cite{bicchi2000robotic,saxena2008robotic, redmon2016you}. As the need for more sophisticated perception increased, depth cameras were introduced to enhance 3D geometry detection, allowing robots to estimate objects' volume and spatial orientation more accurately \cite{redmon2015real,gou2021rgb,fang2023anygrasp}. To further improve perception, tactile sensors were integrated into robotic systems, allowing robots to perceive additional properties, such as force \cite{bhattacharjee2012haptic,lee2024regrasping,chen2012hand}, temperature \cite{decherchi2011tactile} and material stiffness \cite{sinapov2011vibrotactile}. These multimodel sensory inputs greatly improved the robustness of robotic grasping by offering multi-dimensional information. Despite these advances, one critical limitation remains: the inability to distinguish materials with similar visual and physical properties. This gap highlights the need for more advanced sensing technologies that can incorporate additional information modalities, such as spectral data, to enable robots to achieve reliable material differentiation.

\subsection{Hyperspectral Technologies}
Hyperspectral technology captures detailed spectral information across dozens to hundreds of contiguous bands, enabling precise material characterization and analysis \cite{lodhi2019hyperspectral}. This capability has been extensively used in fields, including biochemistry analysis, quality control within the pharmaceutical industry \cite{silva2018hyperspectral, fan2016applications}, non-destructive food inspection \cite{feng2012application}, and industrial sorting \cite{tatzer2005industrial}. Beyond its spectral resolution, hyperspectral images also provide spatial data, enabling pixel-by-pixel analysis, which is invaluable for applications requiring pixel-level analysis, such as environmental mapping and industrial sorting \cite{bioucas2013hyperspectral, tatzer2005industrial}. The adoption of hyperspectral imaging in robotics has been slow due to deployment challenges. Traditional point spectrometers are effective for material analysis but lack spatial context, whereas linescan hyperspectral cameras require complex auxiliary mechanisms for operation. Meanwhile, snapshot hyperspectral cameras usually offer fewer spectral bands at significantly higher costs \cite{thomas2025trends}. Furthermore, the high-dimensional nature of hyperspectral data demands significant computational resources, often impeding its use in real-time robotic applications \cite{makarenko2022real}.

\subsection{Applications of Hyperspectral Technologies in Robotics}
Serval studies have explored the integration of hyperspectral technology with robotics systems. One approach involves embedding spectrometers in robotic end-effectors for material identification. For example, spectrometer-equipped grippers have been used to assess food quality, classify household materials and liquid classification in container \cite{cortes2017integration,erickson2019classification, erickson2020multimodal, hanson2023slurp,li2024m, hanson2024prospect}. Some researchers have also attempted to enhance recognition capabilities in robotic grasping operations by integrating spectrometers into the gripper \cite{hanson2022pregrasp}. While these methods leverage spectral data, they often neglect the spatial information that could further enhance robotic perception.

Another line of research explores using robots to facilitate hyperspectral data acquisition. For instance, Wang et al. employed drones for remote sensing tasks \cite{wang2021combining}, Azizi et al. used the precise movement of robots to carry objects under hyperspectral sensors for data collection \cite{azizi2024autonomous}. And Hanson et al. developed an unmanned ground vehicle (UGV) equipped with hyperspectral imaging devices \cite{hanson2022vast}; additionally, they also proposed a robotic pushbroom hyperspectral data acquisition system coupled with an effective algorithm for detecting occluded hyperspectral objects \cite{hanson2022occluded}. While these approaches leverage both the spectral and spatial information from hyperspectral images, they do not integrate this information into robotic manipulation or task execution.

To address these limitations, our proposed system leverages both spectral and spatial information to enable hyperspectral imaging-guided robotic grasping. By combining a custom-designed hyperspectral device, PRISM, with the SpectralGrasp framework, our work aims to bridge the gap between hyperspectral imaging and robotic manipulation, offering an effective solution for challenging material recognition and grasping tasks in close-range robotic scenarios.

\begin{figure*}[tbp]
\centering
	\includegraphics[width=1\linewidth]{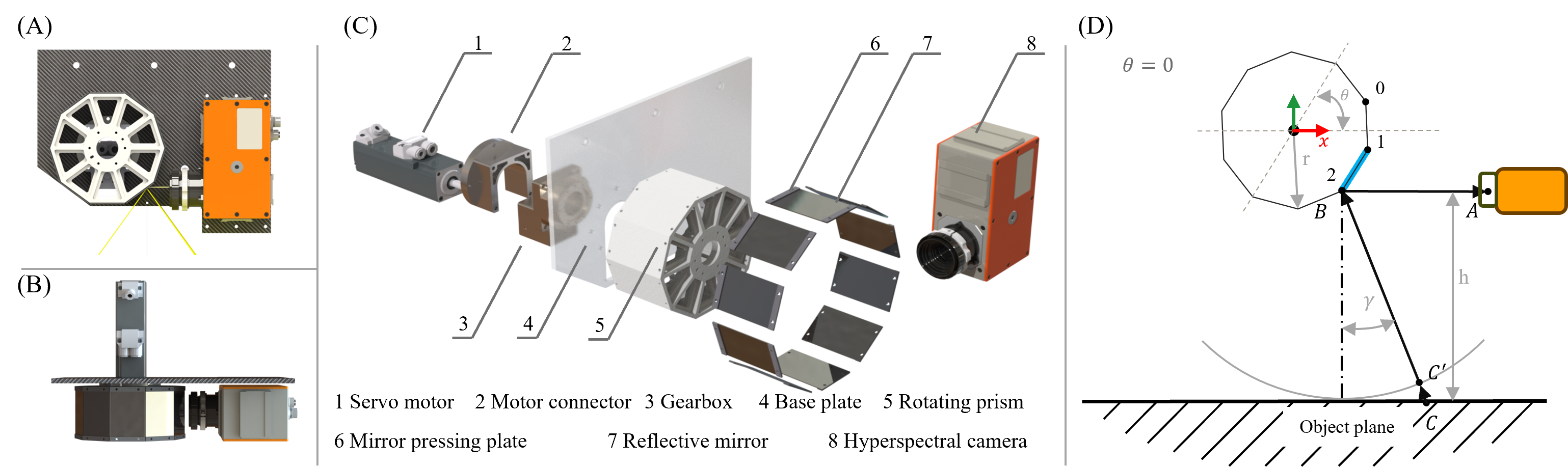}
         \vspace{1pt}
	\caption{CAD renderings of the mechanical design and the working mechanism of PRISM. (A) Fully assembled front view of PRISM, showing the two limited scanning positions. (B) Fully assembled top view. (C) Exploded view of PRISM, illustrating the individual components. (D) The working mechanism of RPISM, highlighting the scanning process. The angle $\theta$ denotes the rotation angle of the servo motor, while the angle $\gamma$ denotes the scanned angle. The circumradius of the rotating prism is $r$, and the position of the hyperspectral sensor, point $A$, is given by $(x_A, \frac{\sqrt{2}r}{2})$.}
    \label{fig::PRISM}

\end{figure*}
\section{METHOD}
The section details the design and implementation of the hyperspectral imaging device PRSIM and the SpectralGrasp framework. These components integrate hyperspectral data into robotic systems by effectively leveraging spectral and spatial information to enhance material recognition and grasping performance. The proposed methodology highlights PRISM's innovative design for distortion-free hyperspectral imaging and demonstrates how SpectralGrasp generates robust grasping strategies from hyperspectral data, as illustrated in Fig. \ref{fig: System Overview}.

\subsection{PRISM}

\subsubsection{Design}
The PRISM (Polyhedral Reflective Imaging Scanning Mechanism) is a specialized hyperspectral imaging device optimized for robotic applications. PRISM comprises three primary components: a high-precision servo motor, a regular decagonal reflective prism, and a linescan hyperspectral camera, as illustrated in Fig. \ref{fig::PRISM}. PRISM simplifies the integration and operation of hyperspectral imaging systems by employing rotational reflection, thereby eliminating the need for complex conveyor mechanisms commonly required in traditional line-scan hyperspectral systems. During operation, the servo motor rotates the prism, sequentially reflecting the scanning line area onto the sensor of the hyperspectral camera via a reflective mirror. Compared to commercial snapshot hyperspectral cameras of equivalent spectral range, PRISM provides higher spectral resolution and precision at significantly lower costs. 

The servo motor achieves control accuracy of 0.01 degrees, augmented by a high-precision encoder for real-time positional feedback. A gearbox with a reduction ratio of 10 enhances rotational accuracy and reduces torque requirements. Each facet of the decagonal reflective prism features high-reflectivity silvered mirrors, reflecting approximately 95\% of incident light back onto the hyperspectral camera’s sensor. 

The number of sides of the reflective prism directly influences the PRSIM's field of view (FOV). More sides result in a narrower scanning area but higher scanning precision. The relationship between the FOV and the number of the reflective prism sides $n$ is defined by the following formula: 
\begin{equation} 
   FOV = \frac{720\degree}{n}
\end{equation}

A decagonal prism configuration ensures uniform reflection and consistent imaging quality, providing an optimal balance between range and precision, yielding a practical FOV of 72\degree.

\subsubsection{Hyperspectral Image Reconstruction}

As shown in Fig. \ref{fig::PRISM} (D), in the PRISM configuration, the scanned angle $\gamma$ is a linear function of the motor angle $\theta$, meaning that for each motor position during the scanning cycle, there is a corresponding unique scanning position. The relationship between the motor angle $\theta$ and the scanned angle $\gamma$ is:
\begin{equation}
   \gamma = \frac{3\pi}{10} - 2\theta,   \theta \in [\frac{\pi}{20},\frac{\pi}{4}]
\label{equ:gamma}
\end{equation}

The operational range defined by $\gamma$ ranges from $\gamma \in  [-\frac{\pi}{5},\frac{\pi}{5}] $ facilitates uniform spatial resolution across hyperspectral images. Spatial resolution of PRISM hyperspectral images measures $H \times W = 871 \times 512$ pixels,  $N$ representing the number of spectral bands. Each vertical direction position $h$ corresponds to a unique motor angle $\theta_h$. During the imaging acquisition, each captured frame $f \in \mathbb{R} ^ {1\times W\times N}$ is recorded along with its corresponding motor angle $\theta_h$. These frames are then accurately mapped to their respective positions in the sample image based on the recorded angles, ultimately forming the complete hyperspectral image $I \in \mathbb{R} ^ {H\times W\times N}$. Typically, the motor is driven at 3 rpm. yielding a full-scan duration of approximately 2s.

\subsubsection{Distortion Correction}

\begin{figure}[tbp!]
\centering
	\includegraphics[width=1\linewidth]{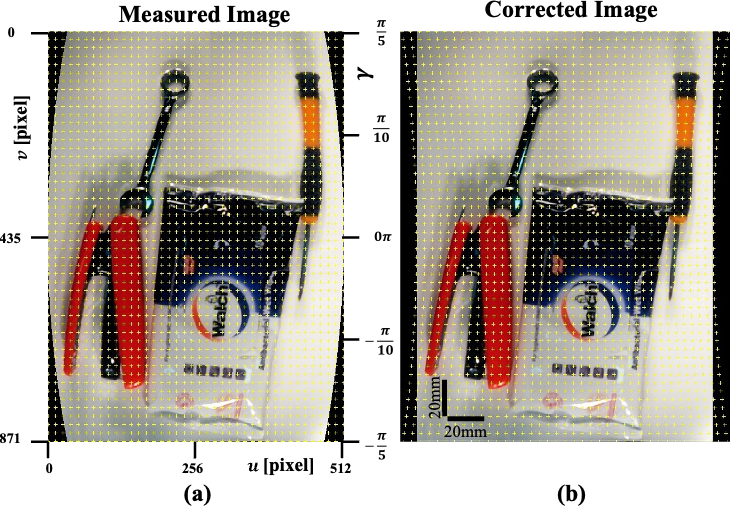}
        \vspace{1pt}
	\caption{Curvature distortion correction. (a) Original image exhibiting distortion with increased scanned angle $\gamma$. (b) Corrected image displaying uniform spatial mapping in both the $x$ and $y$ directions.}
    \label{fig:: distortion}
\end{figure}

Reflective mirrors introduce curvature distortion in the field of view (FOV). This distortion becomes increasingly pronounced as the scanned angle $\gamma$ increases, visibly warping captured images, as illustrated in Fig. \ref{fig:: distortion}. Distortion in object width per scan line follows scaling factor $k(\theta)$, which is a nonlinear function of motor angle $\theta$: 
\begin{align}
    k(\theta) &= \sqrt{1+ \frac{1}{\tan^2(\frac{\pi}{5}+2\theta)}} 
\end{align}

Each pixel coordinate $(u,v)$ in the hyperspectral image $I$ corresponds to specific spatial coordinates $(x_d, y_d)$, the relationship is described by:
\begin{align}
    y_d &= h* tan(\gamma)\\
    x_d &= u* \triangle x *k(\theta)
\end{align}

where $\triangle x$ denotes the spatial resolution of each measured line in millimeters, determined by the camera lens properties and detection height $h$. To correct the curvature distortion, a transformation matrix $M(\theta)$ is introduced incorporating the scaling factor $k(\theta)$ to accurately map each pixel's spatial coordinates, resulting in the corrected hyperspectral image $I^{c}$.
\begin{align}
   I^{c} = M(\theta(u,v)) \cdot I(u,v)
\end{align}
The complete nonlinear transformation effectively removes distortion, as demonstrated in Fig. \ref{fig:: distortion} (b). Although distortion has minimal impact on grasp estimation for non-detailed objects, the correction was utilized in the following experiments for optimal system performance.

\subsection{SpectralGrasp}
\begin{figure}[t] %
    \centering
    \includegraphics[width=0.6\linewidth]{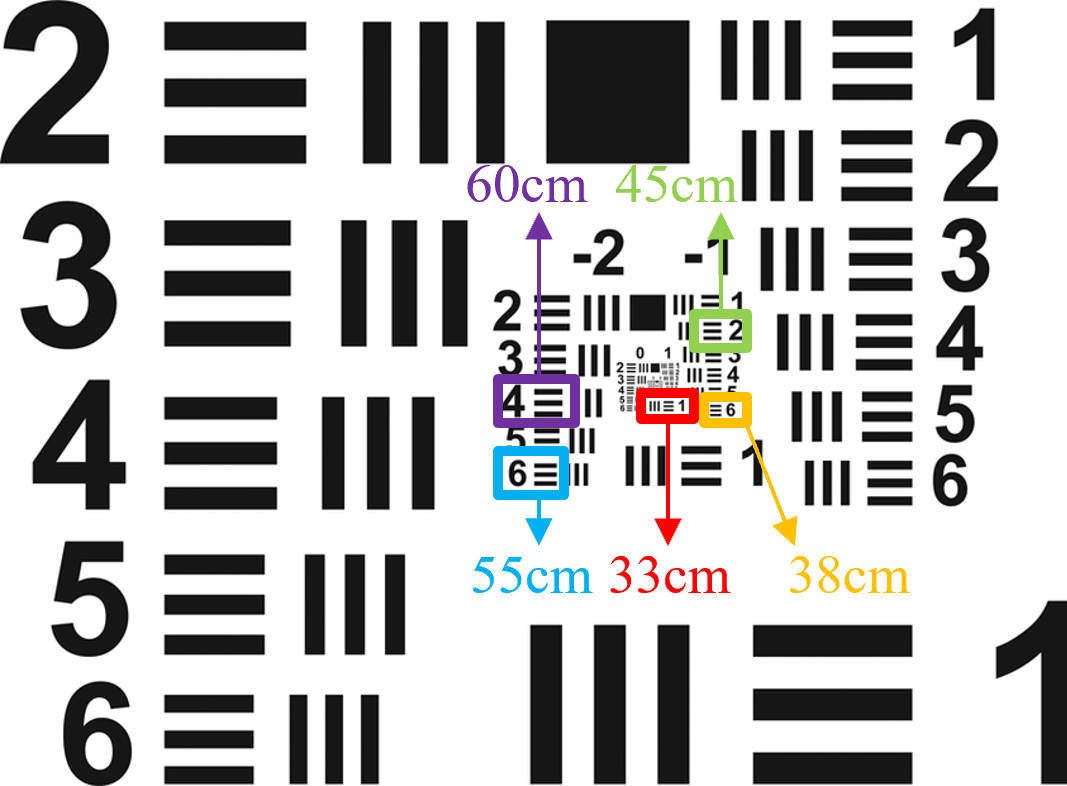} 
    \small
    
    \centering
    \vspace{1em}
    \begin{tabular}{| c | c | c | c | c |c | c |}
        \hline
         Working Height (mm) & 330 & 380 & 450 & 550 & 600\\
        \hline
        Resolution (lp/mm)& 4& 3.56 & 2.24 & 1.78 & 1.41\\
        \hline
        Min Resolvable Size (mm)& 0.13& 0.14 & 0.22 & 0.28 & 0.35\\
        \hline
    \end{tabular}
    \vspace{1em}
    \caption{ Spatial resolution test results for PRISM at various experimental heights. The resolution was measured using the 1951 USAF Resolution Test Chart (MIL-STD-150), containing reference line patterns with known dimensions. The diagram illustrates the achieved resolution in line pairs per millimeter (lp/mm) at heights ranging from 33 cm to 60 cm. The table below summarizes the spatial resolution performance and minimum resolvable size.} 
    \label{fig:: spatial_resolution}
\end{figure}
\subsubsection{Overview}
This section introduces SpectralGrasp, a framework designed to utilize spectral and spatial information from hyperspectral images obtained by PRISM to generate robotic grasping strategies.  

\subsubsection{Hyperspectral Imaging Preprocess}
As previously discussed, the substantial volume of hyperspectral data presents a significant challenge for real-time processing, given its complexity relative to standard RGB imagery. Identifying and isolating the region of interest (ROI) of target objects is therefore critical for enhancing real-time processing capabilities.

Initially, spatial object information is extracted from the corrected hyperspectral image $I^{c}$. Segmentation masks $M \in {0,1}^{H \times W}$ for each object are generated using the Segment Anything model \cite{kirillov2023segment} applied to pseudo-RGB images derived from $I^{c}$. These masks, combined with their respective hyperspectral data, produce masked hyperspectral images $I^{M}$, which proceed to further classification and localization steps. Redundant spectral information in hyperspectral images can negatively affect detection accuracy and processing speed; thus, a minimum noise fraction (MNF) method is applied to retain the most informative frequency bands \cite{lee1990enhancement}. The MNF method, implemented through the Python spectral library (spy) , typically compresses around thirty percent of spectral bands to optimize processing speed and recognition accuracy, effectively reducing computational load while preserving essential spectral information.

\subsubsection{Object Detection and Grasp Point Prediction}
\label{sec:: grasping point}
The objective is to classify each object and identify appropriate suction points. Traditional RGB-based detection models are inadequate due to hyperspectral image dimensionality. Thus, a specialized two-stage neural network for hyperspectral object detection is developed.

In the initial stage, pixel-level classification occurs within the masked hyperspectral image $I^{M}$. The spectral data of each pixel $I^{M}(u,v)$ serve as inputs for the network, initially processed by one-dimensional convolution (Conv1d) layers to establish feature channels. Subsequently, one-dimensional max pooling (MaxPool1d) reduces spectral vector lengths for frequency-domain downsampling, followed by a ReLU activation function to introduce nonlinearity and batch normalization to standardize intermediate features. Iterating this convolution–pooling–normalization sequence reduces hyperspectral data dimensionality, resulting in compact feature representations, which are subsequently flattened and classified using a fully connected layer (nn.Linear). 

The second stage aggregates pixel-level classifications into object-level classifications through Principal Component Analysis (PCA) clustering \cite{rodarmel2002principal}. Geometric centroids of identified objects serve as candidate suction points. PCA's selection reflects its proven effectiveness for hyperspectral classification scenarios, though alternative methods such as Support Vector Machines (SVM) could also be implemented.

\subsubsection{Trajectory Planning}
A waypoint-based trajectory planner generates robot trajectories.  Based on suction points and target bin locations identified from object classes, a sparse trajectory is initially established using predefined intermediate waypoints. This sparse trajectory undergoes refinement through Linear Quadratic Tracking \cite{modares2014linear}, producing smooth Cartesian trajectories at 100 Hz for subsequent execution by the robot's Cartesian controller.

\begin{figure*}[t]
\centering
	\includegraphics[width=1\linewidth]{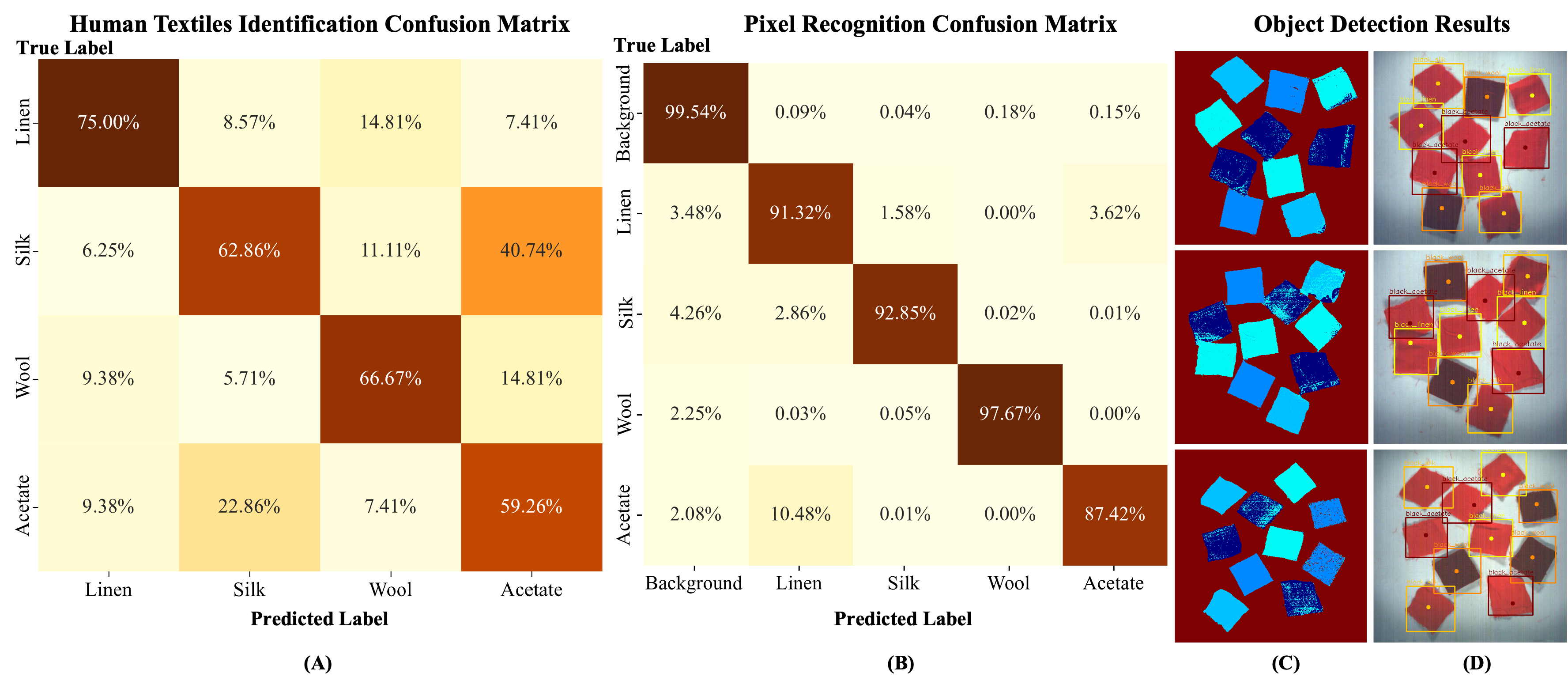}
	\caption{Textile Recognition Experiment Results. Textile recognition experiment results. The comparison between human identification (A) and the system's pixel-level recognition results (B) highlights the significantly higher success rate of the proposed method. The object detection results demonstrate the effectiveness of our algorithm in processing segmentation maps (C) to extract semantic information (D), including object labels, bounding boxes, and suction points for following grasping execution.} 
    \label{fig:: identification result}
\end{figure*}

\section{Experiment}
A series of experiments was conducted to evaluate the performance of the PRISM device and the SpectralGrasp framework. These experiments assessed spatial imaging resolution, material recognition accuracy, and robotic sorting performance in practical scenarios.

\subsection{Spatial Resolution}
PRISM employs a rotational scanning mechanism to eliminate the reliance on conveyor belts commonly required by linescan hyperspectral cameras. This design simplifies the deployment of hyperspectral imaging systems, enables the capture of static objects, and broadens the potential application scenario. Exploring the spatial resolution achievable through the rotational scanning function is essential to figure out the imaging performance of PRISM.

Spatial resolution, defined as the smallest distinguishable feature size, was quantified using the 1951 USAF Resolution Test Chart (MIL-STD-150), which contains standardized reference patterns. Tests at multiple heights ranging from 330 mm to 600 mm were performed, and the outcomes are illustrated as Fig. \ref{fig:: spatial_resolution}. The results demonstrate that PRISM can achieve a minimum resolvable size of 0.35mm at a working height of 600mm, which is sufficient for guiding close-range robotic perception and grasping tasks.

\subsection{Textile Recognition Experiment}
PRISM's recognition performance and our algorithm's accuracy were evaluated by distinguishing various textile materials of identical color. Additionally, the effectiveness of PRISM was benchmarked against human performance, reflecting typical conditions in textile recycling factories, where fabric sorting often relies heavily on human visual inspection. Four textile materials—linen, wool, acetate, and silk—each prepared in black, white, blue, and yellow colors, comprised the experimental dataset. Four trained participants independently performed textile identification tasks with mixed piles containing 16 distinct samples across four repetitions.

Results in Fig. \ref{fig:: identification result} (A) showed an average human identification accuracy of 66\%. In contrast, the pixel-level recognition classification results achieved accuracy ranging from 87\% to frequently surpassing 91\%, as illustrated in Fig. \ref{fig:: identification result} (B). Object-level classification results further demonstrated robustness, as occasional pixel-level errors did not substantially influence object recognition accuracy or grasping point determination, as shown in Fig. \ref{fig:: identification result} (C) and Fig. \ref{fig:: identification result} (D). Overall, our algorithm robustly recognizes various materials and provides reliable grasping information, effectively leveraging hyperspectral data to guide robotic manipulation tasks.

\begin{figure}[t]
\centering
	\includegraphics[width=1\linewidth]{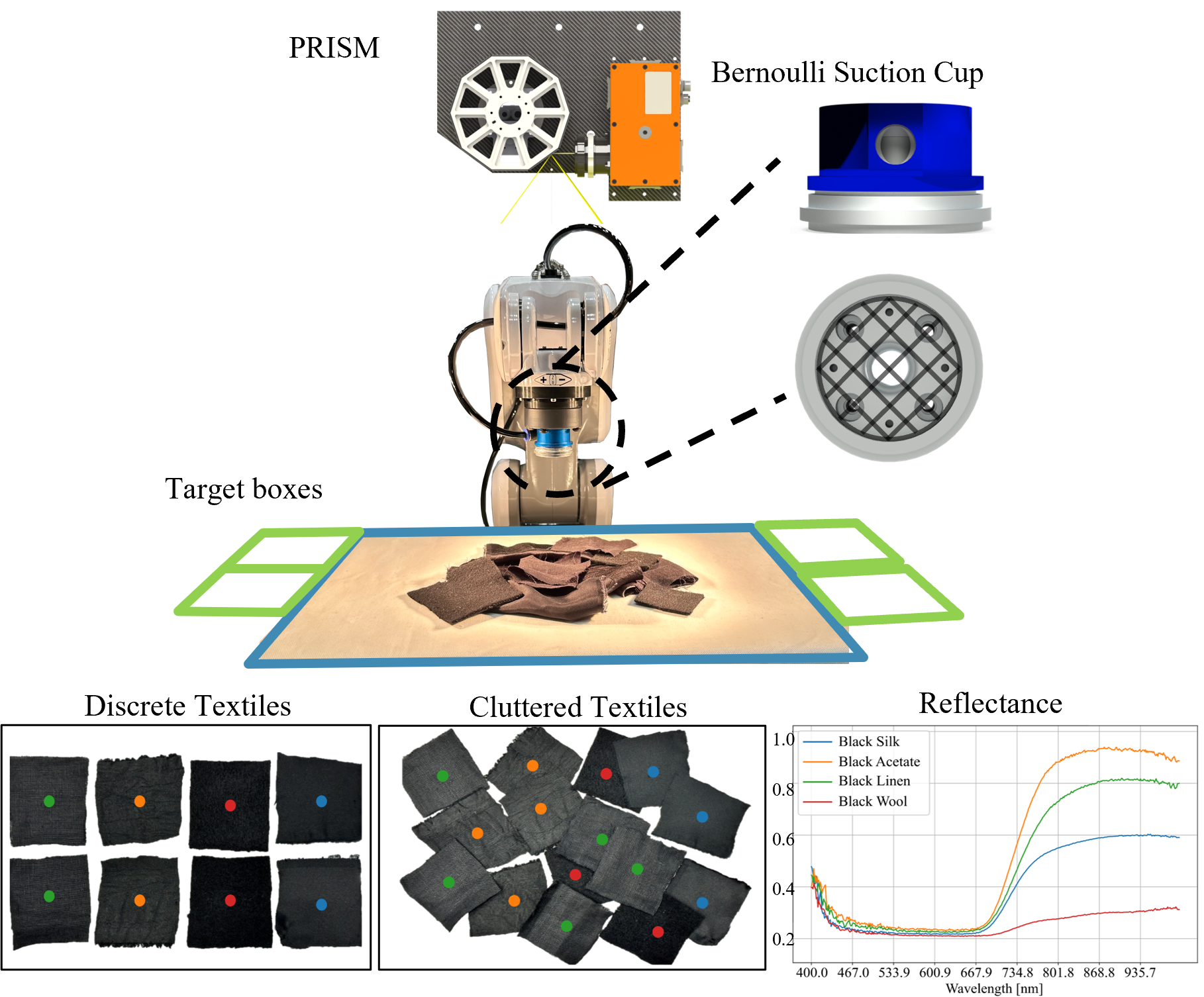}
         \vspace{1pt}
	\caption{Experiment setup for textile sorting experiment. Textiles are sorted from the object plane (surrounded by blue lines) to the corresponding target boxes (surrounded by green lines). Hyperspectral images are captured by the overhead PRISM system. We install a screen net to the Bernoulli suction cup's outlet to ensure effective suction. Performance is evaluated on two difficult levels black textiles, including silk, acetate, linen, and wool. Their spectrum curves are illustrated.} 
    \label{fig:: Textile experiment}
\end{figure}

\subsection{Hyperspectral Guided Sorting Experiment}
A sorting experiment was conducted to verify the efficacy of PRISM and SpectralGrasp in real-world robotic sorting applications. The primary objective of this experiment is to demonstrate the efficacy of our hyperspectral imaging-guided robotic grasping system in accurately generating grasping points based on spectral-spatial information. Textile sorting is selected as the validation scenario because it effectively highlights the advantages of hyperspectral imaging in differentiating materials and object contours at pre-grasp stage, thereby enhancing grasping reliability.

\subsubsection{Setup}
For the hyperspectral guided sorting experiment, a NACHI MZ04 6-axis robotic arm was equipped with a Bernoulli suction cup gripper. This specialized suction cup enables the lifting of unsealed textile pieces without disturbing the surrounding items. The suction cup is powered by a high-pressure airflow and controlled via a relay connected to a Modbus 485 module, ensuring consistent and reliable gripping performance. 
The PRISM is mounted above the workspace to capture global scene information. Four designated collection boxes are positioned around the workspace to receive the sorted textiles, as illustrated in Fig. \ref{fig:: Textile experiment}. All experiments are conducted on a desktop computer running Windows 11, equipped with an Intel i7-13700KF CPU and an NVIDIA RTX 4090 GPU. The core codebase is implemented in C++ for computational efficiency, while additional Python scripts are utilized to simplify system interfacing. 

\begin{algorithm}[]
\caption{Textiles Sorting Experiments}\label{algorithm: grasping}
\begin{algorithmic}[1]
\State \textbf{Initialize:} robot.ready(), suction\_cup.close()
\State \textbf{Initialize:} object\_present $\gets$ \textbf{true}, target\_bin

\While{object\_present}
    \State \textcolor{blue}{\LeftComment{Stage 1: Capture and process hyperspectral image}}
    \State $\mathcal{I}^c \gets$ PRISM.capture().preprocess()
    \State $\mathcal{M} \gets$ SegmentObjects($\mathcal{I}^c.rgb$)

    \If{$\mathcal{M}$ is empty} 
        \State object\_present $\gets$ \textbf{false}
        \State \textbf{break}
    \EndIf

    \State \textcolor{blue}{\LeftComment{Stage 2: SpectralGrasp}}
    \State $\mathcal{L}_{pixels} \gets$ PredictPixelLabels($\mathcal{I}^c,{M}$)
    \State $objects \gets$ PCA($\mathcal{L}_{pixels}$)
    \State \textcolor{blue}{\LeftComment{Stage 3: Grasp pose calculation and execution}}
    \For{ object $\in$ objects}
        \State $\mathcal{G} \gets$ GetGraspPoses(object)
        \State \textbf{robot}.move($\mathcal{G}$), \textbf{suction\_cup}.open()
        \State \textbf{robot}.move(target\_bin[object])
        \State \textbf{suction\_cup}.close()
    \EndFor
\EndWhile
\end{algorithmic}
\end{algorithm}

\subsubsection{Experimental Procedure}
The experiment aimed to recognize textiles and accurately sort them into the corresponding collection bins using the hyperspectral images obtained from PRISM and processed by the SpectralGrasp algorithm. The system's capability was assessed using four textile types: Linen, Silk, Wool, and Acetate, as shown in Fig. \ref{fig:: Textile experiment}. The spectral reflectances of these textiles are also shown in Fig. \ref{fig:: Textile experiment}. Two experimental conditions were designed to evaluate performance under varied complexities:
\begin{enumerate}[1)] 
\item Discrete – Textiles are placed separately without overlap, allowing clear delineation of individual items.
\item Cluttered – Textiles are randomly arranged with overlaps, posing significant challenges in distinguishing boundaries.
\end{enumerate}

Although previous comparisons with human identification demonstrated superior textile recognition capability of our proposed system, A comparative analysis against a state-of-the-art RGB-based detection method, YOLOc11 \cite{redmon2016you}, was also performed. This approach allowed a direct comparison of the sorting performance between RGB-guided sorting and our hyperspectral-guided approach, highlighting the value of spectral-spatial information.

52 individual samples were selected from four different types of textiles as the test set, with each sample typically measuring approximately 5cm $\times$ 5cm. In each trial, textiles were introduced into the sorting area according to the defined scenario, with minimal manual intervention. The complete experimental procedure is summarized in Algorithm \ref{algorithm: grasping}.

\subsubsection{Results}
\begin{figure}[]
\vspace{5pt}
\centering
	\includegraphics[width=1\linewidth]{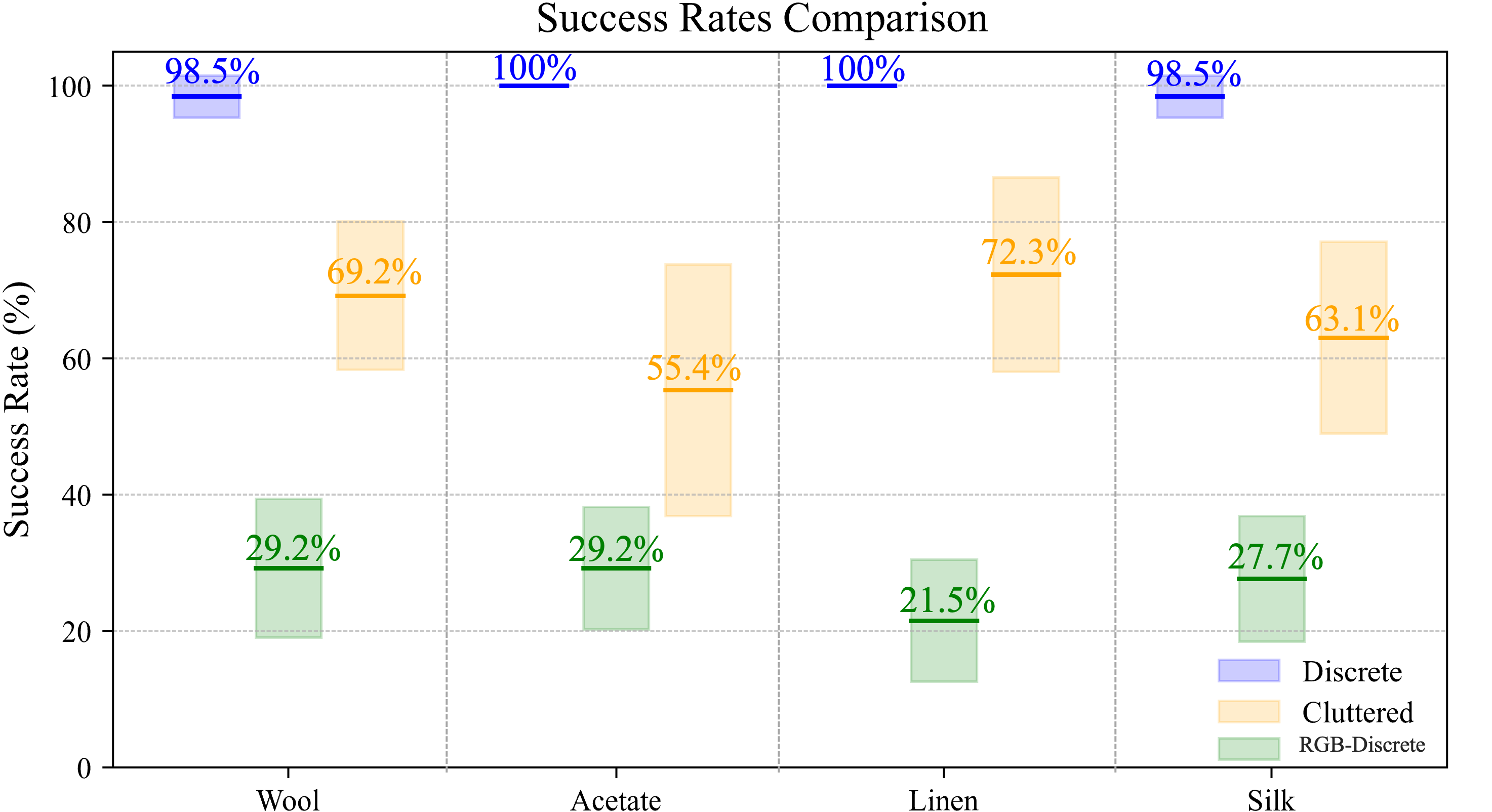}
    \vspace{1pt}
	\caption{Comparison of sorting success rates under three experimental conditions—Discrete, Cluttered and RGB guided Discrete —for four types of black textiles: Wool, Acetate, Linen, and Silk. The shaded regions indicate the mean ± one standard deviation of the success rates, while the horizontal lines within these shaded areas mark the mean values.} 
    \label{fig:: sorting results}
\end{figure}

 Performance results from five independent trials across the discrete hyperspectral, cluttered hyperspectral, and discrete RGB scenarios are summarized in Fig. \ref{fig:: sorting results}.Under discrete conditions, hyperspectral-guided sorting achieved near-perfect performance. In contrast, RGB-guided sorting exhibited significantly lower effectiveness. The results strongly support hyperspectral imaging’s role in enhancing robotic grasping and sorting capabilities, particularly under scenarios challenging for traditional RGB methods. However, when transitioning from the relatively simple discrete scenario to the more complex cluttered scenario, sorting accuracy declined by 27\% to 45\% across the four textiles, with increased variability (standard deviation) in the results, demonstrating that pure hyperspectral sensing performance degrades in complex scenes. These findings do not imply that hyperspectral imaging universally outperforms RGB vision; rather, they underscore its value as a complementary sensing modality capable of significantly enhancing robotic perception and grasping capabilities in challenging environments.

\subsection{Comparative Experiments}

To the best of our knowledge, few studies have specifically addressed hyperspectral imaging-guided robotic grasping tasks, making it challenging to directly compare the grasp-point prediction accuracy of our proposed system with other existing methods. An indirect evaluation was therefore carried out by substituting the pixel‑level classification module with alternative hyperspectral detection algorithms and measuring both recognition accuracy and computational efficiency. Benchmarks included the 3D‑CNN approaches of Mei et al. \cite{mei2019unsupervised} and Li et al. \cite{li2017spectral}, as well as the 1D semi‑supervised method of Boulch et al. \cite{boulch2017autoencodeurs}. Results in Table \ref{tab:comparison-results} show that the proposed network reduces training time to 1.41h and inference time to 21.0s per image—substantially lower than the 84h and 785s reported by Mei et al. and the 2.85h and 27.6s reported by Li et al.—making it far better suited for real‑time grasping and manipulation scenarios. Accuracy remains competitive at 98.02\%, differing by less than 0.7\% from Li et al. (98.52\%) and Boulch et al. (98.67\%). Because most existing methods were developed for large‑scale remote‐sensing applications, where real‑time performance is less critical, this balance of speed and accuracy underscores the suitability of the proposed classifier for object‑level robotic tasks. The framework also allows other hyperspectral classifiers to be integrated as needed.

\section{CONCLUSION AND DISCUSSION}
\begin{table}[t]
\centering
\caption{Comparative Experiment Results}
\label{tab:comparison-results}
\resizebox{\columnwidth}{!}{%
\begin{tabular}{lcccc}
\hline
\textbf{Evaluation Parameter} & \multicolumn{4}{c}{\textbf{Hyperspectral Recognition}} \\
\cline{2-5}
& \textbf{Ours} & Mei et al. \cite{mei2019unsupervised} & Li et al. \cite{li2017spectral} & boulch et al.\cite{boulch2017autoencodeurs}  \\
\hline
Training time (/h)             & \textbf{1.41} & 84 & 2.85 & 1.52  \\
Inference time per image(/sec)     & \textbf{21.0} & 785 & 27.6 & 22.1 \\
Accuracy (/\%) & 98.02 & 96.6 & 98.52 & \textbf{98.67}  \\
\hline
\end{tabular}
}
\end{table}

 This study introduced a hyperspectral imaging-guided robotic grasping system, comprising PRISM and the SpectralGrasp framework, to enhance robotic perception and grasping accuracy. PRISM provides high-precision, distortion-free imaging with streamlined integration, while SpectralGrasp leverages spectral-spatial hyperspectral information to generate robust grasping strategies. Experiments involving textile recognition and sorting validated the system’s superior performance compared to human operators and RGB-based methods. Results emphasized the capability of hyperspectral imaging as a valuable perception modality in robotics. Future research directions include extending the proposed approach to more dynamic and complex manipulation scenarios. Additionally, integration of complementary sensing modalities, such as geometric or depth information, will be explored to further improve the robustness and versatility in diverse robotic grasping and manipulation tasks.

\balance 
\bibliographystyle{IEEEtran}
\bibliography{main}

@article{lodhi2019hyperspectral,
  title={Hyperspectral imaging system: Development aspects and recent trends},
  author={Lodhi, Vaibhav and Chakravarty, Debashish and Mitra, Pabitra},
  journal={Sensing and Imaging},
  volume={20},
  pages={1--24},
  year={2019},
  publisher={Springer}
}

@article{saxena2008robotic,
  title={Robotic grasping of novel objects using vision},
  author={Saxena, Ashutosh and Driemeyer, Justin and Ng, Andrew Y},
  journal={The International Journal of Robotics Research},
  volume={27},
  number={2},
  pages={157--173},
  year={2008},
  publisher={Sage Publications Sage UK: London, England}
}

@inproceedings{redmon2015real,
  title={Real-time grasp detection using convolutional neural networks},
  author={Redmon, Joseph and Angelova, Anelia},
  booktitle={2015 IEEE international conference on robotics and automation (ICRA)},
  pages={1316--1322},
  year={2015},
  organization={IEEE}
}

@inproceedings{bicchi2000robotic,
  title={Robotic grasping and contact: A review},
  author={Bicchi, Antonio and Kumar, Vijay},
  booktitle={Proceedings 2000 ICRA. Millennium conference. IEEE international conference on robotics and automation. Symposia proceedings (Cat. No. 00CH37065)},
  volume={1},
  pages={348--353},
  year={2000},
  organization={IEEE}
}

@article{fang2023anygrasp,
  title={Anygrasp: Robust and efficient grasp perception in spatial and temporal domains},
  author={Fang, HaoShu and Wang, Chenxi and Fang, Hongjie and Gou, Minghao and Liu, Jirong and Yan, Hengxu and Liu, Wenhai and Xie, Yichen and Lu, Cewu},
  journal={IEEE Transactions on Robotics},
  year={2023},
  publisher={IEEE}
}

@article{bioucas2013hyperspectral,
  title={Hyperspectral remote sensing data analysis and future challenges},
  author={Bioucas-Dias, JosM and Plaza, Antonio and Camps-Valls, Gustavo and Scheunders, Paul and Nasrabadi, Nasser and Chanussot, Jocelyn},
  journal={IEEE Geoscience and remote sensing magazine},
  volume={1},
  number={2},
  pages={6--36},
  year={2013},
  publisher={IEEE}
}

@article{erickson2019classification,
  title={Classification of household materials via spectroscopy},
  author={Erickson, Zackory and Luskey, Nathan and Chernova, Sonia and Kemp, Charles C},
  journal={IEEE Robotics and Automation Letters},
  volume={4},
  number={2},
  pages={700--707},
  year={2019},
  publisher={IEEE}
}

@inproceedings{erickson2020multimodal,
  title={Multimodal material classification for robots using spectroscopy and high resolution texture imaging},
  author={Erickson, Zackory and Xing, Eliot and Srirangam, Bharat and Chernova, Sonia and Kemp, Charles C},
  booktitle={2020 IEEE/RSJ International Conference on Intelligent Robots and Systems (IROS)},
  pages={10452--10459},
  year={2020},
  organization={IEEE}
}

@inproceedings{hanson2023slurp,
  title={Slurp! spectroscopy of liquids using robot pre-touch sensing},
  author={Hanson, Nathaniel and Lewis, Wesley and Puthuveetil, Kavya and Furline, Donelle and Padmanabha, Akhil and Padir, Ta{\c{s}}lan and Erickson, Zackory},
  booktitle={2023 IEEE International Conference on Robotics and Automation (ICRA)},
  pages={3786--3792},
  year={2023},
  organization={IEEE}
}

@inproceedings{lee2024regrasping,
  title={Regrasping on Printed Circuit Boards with the Smart Suction Cup},
  author={Lee, Jungpyo and Sun, Zheng and Dong, Zhipeng and Chen, Fei and Stuart, Hannah S},
  booktitle={2024 IEEE International Conference on Robotics and Automation (ICRA)},
  pages={6477--6483},
  year={2024},
  organization={IEEE}
}

@inproceedings{redmon2016you,
  title={You only look once: Unified, real-time object detection},
  author={Redmon, Joseph and Divvala, Santosh and Girshick, Ross and Farhadi, Ali},
  booktitle={Proceedings of the IEEE conference on computer vision and pattern recognition},
  pages={779--788},
  year={2016}
}

@inproceedings{gou2021rgb,
  title={Rgb matters: Learning 7-dof grasp poses on monocular rgbd images},
  author={Gou, Minghao and Fang, Hao-Shu and Zhu, Zhanda and Xu, Sheng and Wang, Chenxi and Lu, Cewu},
  booktitle={2021 IEEE International Conference on Robotics and Automation (ICRA)},
  pages={13459--13466},
  year={2021},
  organization={IEEE}
}

@inproceedings{bhattacharjee2012haptic,
  title={Haptic classification and recognition of objects using a tactile sensing forearm},
  author={Bhattacharjee, Tapomayukh and Rehg, James M and Kemp, Charles C},
  booktitle={2012 IEEE/RSJ International Conference on Intelligent Robots and Systems},
  pages={4090--4097},
  year={2012},
  organization={IEEE}
}

@article{decherchi2011tactile,
  title={Tactile-data classification of contact materials using computational intelligence},
  author={Decherchi, Sergio and Gastaldo, Paolo and Dahiya, Ravinder S and Valle, Maurizio and Zunino, Rodolfo},
  journal={IEEE Transactions on Robotics},
  volume={27},
  number={3},
  pages={635--639},
  year={2011},
  publisher={IEEE}
}

@article{sinapov2011vibrotactile,
  title={Vibrotactile recognition and categorization of surfaces by a humanoid robot},
  author={Sinapov, Jivko and Sukhoy, Vladimir and Sahai, Ritika and Stoytchev, Alexander},
  journal={IEEE Transactions on Robotics},
  volume={27},
  number={3},
  pages={488--497},
  year={2011},
  publisher={IEEE}
}

@article{silva2018hyperspectral,
  title={Hyperspectral reflectance as a tool to measure biochemical and physiological traits in wheat},
  author={Silva-Perez, Viridiana and Molero, Gemma and Serbin, Shawn P and Condon, Anthony G and Reynolds, Matthew P and Furbank, Robert T and Evans, John R},
  journal={Journal of Experimental Botany},
  volume={69},
  number={3},
  pages={483--496},
  year={2018},
  publisher={Oxford University Press UK}
}

@article{fan2016applications,
  title={Applications of NMR spectroscopy to systems biochemistry},
  author={Fan, Teresa W-M and Lane, Andrew N},
  journal={Progress in nuclear magnetic resonance spectroscopy},
  volume={92},
  pages={18--53},
  year={2016},
  publisher={Elsevier}
}

@article{feng2012application,
  title={Application of hyperspectral imaging in food safety inspection and control: a review},
  author={Feng, Yao-Ze and Sun, Da-Wen},
  journal={Critical reviews in food science and nutrition},
  volume={52},
  number={11},
  pages={1039--1058},
  year={2012},
  publisher={Taylor \& Francis}
}

@article{tatzer2005industrial,
  title={Industrial application for inline material sorting using hyperspectral imaging in the NIR range},
  author={Tatzer, Petra and Wolf, Markus and Panner, Thomas},
  journal={Real-Time Imaging},
  volume={11},
  number={2},
  pages={99--107},
  year={2005},
  publisher={Elsevier}
}

@article{cortes2017integration,
  title={Integration of simultaneous tactile sensing and visible and near-infrared reflectance spectroscopy in a robot gripper for mango quality assessment},
  author={Cortes, Victoria and Blanes, Carlos and Blasco, Jose and Ortiz, Coral and Aleixos, Nuria and Mellado, Martin and Cubero, Sergio and Talens, Pau},
  journal={Biosystems Engineering},
  volume={162},
  pages={112--123},
  year={2017},
  publisher={Elsevier}
}

@article{hanson2024prospect,
  title={PROSPECT: Precision Robot Spectroscopy Exploration and Characterization Tool},
  author={Hanson, Nathaniel and Lvov, Gary and Rautela, Vedant and Hibbard, Samuel and Holand, Ethan and DiMarzio, Charles and Pad{\i}r, Ta{\c{s}}k{\i}n},
  journal={arXiv preprint arXiv:2403.17232},
  year={2024}
}

@article{azizi2024autonomous,
  title={Autonomous Hyperspectral Characterisation Station: Robot Aided Measuring of Polymer Degradation},
  author={Azizi, Shayan and Asadi, Ehsan and Howard, Shaun and Muir, Benjamin W and O’Shea, Riley and Bab-Hadiashar, Alireza},
  journal={IEEE Transactions on Automation Science and Engineering},
  year={2024},
  publisher={IEEE}
}

@inproceedings{kirillov2023segment,
  title={Segment anything},
  author={Kirillov, Alexander and Mintun, Eric and Ravi, Nikhila and Mao, Hanzi and Rolland, Chloe and Gustafson, Laura and Xiao, Tete and Whitehead, Spencer and Berg, Alexander C and Lo, Wan-Yen and others},
  booktitle={Proceedings of the IEEE/CVF International Conference on Computer Vision},
  pages={4015--4026},
  year={2023}
}

@incollection{elmasry2010principles,
  title={Principles of hyperspectral imaging technology},
  author={ElMasry, Gamal and Sun, Da-Wen},
  booktitle={Hyperspectral imaging for food quality analysis and control},
  pages={3--43},
  year={2010},
  publisher={Elsevier}
}

@inproceedings{makarenko2022real,
  title={Real-time hyperspectral imaging in hardware via trained metasurface encoders},
  author={Makarenko, Maksim and Burguete-Lopez, Arturo and Wang, Qizhou and Getman, Fedor and Giancola, Silvio and Ghanem, Bernard and Fratalocchi, Andrea},
  booktitle={Proceedings of the IEEE/CVF Conference on Computer Vision and Pattern Recognition},
  pages={12692--12702},
  year={2022}
}

@article{zheng2018discrimination,
  title={A discrimination model in waste plastics sorting using NIR hyperspectral imaging system},
  author={Zheng, Yan and Bai, Jiarui and Xu, Jingna and Li, Xiayang and Zhang, Yimin},
  journal={Waste Management},
  volume={72},
  pages={87--98},
  year={2018},
  publisher={Elsevier}
}

@article{li2024m,
  title={M$^{3}$ Tac: A Multispectral Multimodal Visuotactile Sensor with Beyond-Human Sensory Capabilities},
  author={Li, Shoujie and Yu, Haixin and Pan, Guoping and Tang, Huaze and Zhang, Jiawei and Ye, Linqi and Zhang, Xiao-Ping and Ding, Wenbo},
  journal={IEEE Transactions on Robotics},
  year={2024},
  publisher={IEEE}
}

@inproceedings{hanson2022vast,
  title={Vast: Visual and spectral terrain classification in unstructured multi-class environments},
  author={Hanson, Nathaniel and Shaham, Michael and Erdogmus, Deniz and Padir, Taskin},
  booktitle={2022 IEEE/RSJ International Conference on Intelligent Robots and Systems (IROS)},
  pages={3956--3963},
  year={2022},
  organization={IEEE}
}

@article{wang2021combining,
  title={Combining spectral and textural information in UAV hyperspectral images to estimate rice grain yield},
  author={Wang, Fumin and Yi, Qiuxiang and Hu, Jinghui and Xie, Lili and Yao, Xiaoping and Xu, Tianyue and Zheng, Jueyi},
  journal={International Journal of Applied Earth Observation and Geoinformation},
  volume={102},
  pages={102397},
  year={2021},
  publisher={Elsevier}
}

@article{rodarmel2002principal,
  title={Principal component analysis for hyperspectral image classification},
  author={Rodarmel, Craig and Shan, Jie},
  journal={Surveying and Land Information Science},
  volume={62},
  number={2},
  pages={115--122},
  year={2002},
  publisher={American Congress on Surveying and Mapping}
}

@article{hanson2022occluded,
  title={Occluded object detection and exposure in cluttered environments with automated hyperspectral anomaly detection},
  author={Hanson, Nathaniel and Lvov, Gary and Padir, Ta{\c{s}}k{\i}n},
  journal={Frontiers in Robotics and AI},
  volume={9},
  pages={982131},
  year={2022},
  publisher={Frontiers Media SA}
}

@article{modares2014linear,
  title={Linear quadratic tracking control of partially-unknown continuous-time systems using reinforcement learning},
  author={Modares, Hamidreza and Lewis, Frank L},
  journal={IEEE Transactions on Automatic control},
  volume={59},
  number={11},
  pages={3051--3056},
  year={2014},
  publisher={IEEE}
}

@article{hanson2022pregrasp,
  title={Pregrasp object material classification by a novel gripper design with integrated spectroscopy},
  author={Hanson, Nathaniel and Kelestemur, Tarik and Erdogmus, Deniz and Padir, Taskin},
  journal={arXiv preprint arXiv:2207.00942},
  year={2022}
}

@article{thomas2025trends,
  title={Trends in Snapshot Spectral Imaging: Systems, Processing, and Quality},
  author={Thomas, Jean-Baptiste and Lapray, Pierre-Jean and Le Moan, Steven},
  journal={Sensors},
  volume={25},
  number={3},
  pages={675},
  year={2025},
  publisher={MDPI}
}

@article{lee1990enhancement,
  title={Enhancement of high spectral resolution remote-sensing data by a noise-adjusted principal components transform},
  author={Lee, James B and Woodyatt, A Stephen and Berman, Mark},
  journal={IEEE Transactions on Geoscience and Remote Sensing},
  volume={28},
  number={3},
  pages={295--304},
  year={1990},
  publisher={IEEE}
}

@inproceedings{boulch2017autoencodeurs,
  title={Autoencodeurs pour la visualisation d’images hyperspectrales},
  author={Boulch, A and Audebert, N and Dubucq, D},
  booktitle={Proc. 25th Colloque Gretsi},
  pages={1--4},
  year={2017}
}

@article{li2017spectral,
  title={Spectral--spatial classification of hyperspectral imagery with 3D convolutional neural network},
  author={Li, Ying and Zhang, Haokui and Shen, Qiang},
  journal={Remote Sensing},
  volume={9},
  number={1},
  pages={67},
  year={2017},
  publisher={MDPI}
}

@article{mei2019unsupervised,
  title={Unsupervised spatial--spectral feature learning by 3D convolutional autoencoder for hyperspectral classification},
  author={Mei, Shaohui and Ji, Jingyu and Geng, Yunhao and Zhang, Zhi and Li, Xu and Du, Qian},
  journal={IEEE Transactions on Geoscience and Remote Sensing},
  volume={57},
  number={9},
  pages={6808--6820},
  year={2019},
  publisher={IEEE}
}

@inproceedings{chen2012hand,
  title={i-hand: An intelligent robotic hand for fast and accurate assembly in electronic manufacturing},
  author={Chen, Fei and Sekiyama, Kosuke and Di, Pei and Huang, Jian and Fukuda, Toshio},
  booktitle={2012 IEEE International Conference on Robotics and Automation},
  pages={1976--1981},
  year={2012},
  organization={IEEE}
}

@inproceedings{sun2024prism,
  title={PRISM: A Polyhedral Reflective Imaging Scanning Mechanism for Hyperspectral Applications},
  author={Sun, Zheng and Dong, Zhipeng and Wang, Shixiong and Chen, Fei},
  booktitle={2024 IEEE International Conference on Robotics and Biomimetics (ROBIO)},
  pages={1353--1358},
  year={2024},
  organization={IEEE}
}
\end{document}